\documentclass[conference]{IEEEtran}
\IEEEoverridecommandlockouts
\usepackage{hyperref}
\usepackage[T1]{fontenc}
\usepackage{graphicx}
\usepackage{subfigure}
\usepackage{setspace}
\usepackage{xcolor}
\usepackage{multicol}
\DeclareUnicodeCharacter{2212}{-}
\DeclareUnicodeCharacter{2009}{-}
\usepackage{amsmath}
\usepackage{amssymb}
\usepackage{float}

\usepackage{caption}
\usepackage{cite}
\usepackage{amsmath,amssymb,amsfonts}
\usepackage{graphicx}
\usepackage{array}
\usepackage{booktabs}
\usepackage{caption}
\usepackage{ragged2e}
\usepackage{makecell} 
\usepackage{adjustbox} 
\usepackage{anyfontsize} 
\usepackage{booktabs}
\usepackage{array}
\usepackage{tabularx} 
\usepackage{array}
\usepackage{graphicx}
\usepackage{textcomp}
\usepackage{xcolor}
\usepackage{multirow}
\usepackage{booktabs}
\usepackage{graphicx}
\usepackage{multirow}
\usepackage{threeparttable}

\usepackage{xcolor}
\usepackage{cite}
\usepackage{amsmath,amssymb,amsfonts}
\usepackage{algorithmic}
\usepackage{graphicx}
\usepackage{textcomp}
\def\BibTeX{{\rm B\kern-.05em{\sc i\kern-.025em b}\kern-.08em
    T\kern-.1667em\lower.7ex\hbox{E}\kern-.125emX}}
\usepackage{amsmath,amssymb,amsfonts}
\usepackage{graphicx}
\usepackage{textcomp}
\usepackage{xcolor}
\usepackage{multirow}
\usepackage{booktabs}
\usepackage{tcolorbox}
\usepackage{hyperref}
\usepackage{algorithm}
\usepackage{algorithmic}
\usepackage{graphicx}
\usepackage{adjustbox}

\usepackage{xcolor}
\usepackage{cite}
\usepackage{amsmath,amssymb,amsfonts}
\usepackage{algorithmic}
\usepackage{graphicx}
\hypersetup{
	colorlinks=true,
	linkcolor=blue,
	filecolor=magenta,      
	urlcolor=cyan,
	pdftitle={Overleaf Example},
	pdfpagemode=FullScreen,
}
\usepackage{textcomp}
\usepackage{xcolor}
\def\BibTeX{{\rm B\kern-.05em{\sc i\kern-.025em b}\kern-.08em
    T\kern-.1667em\lower.7ex\hbox{E}\kern-.125emX}}

\title{LAET: A Layer-wise Adaptive Ensemble Tuning Framework for Pretrained Language Models}

\author{\IEEEauthorblockN{Jawad Ibn Ahad\textsuperscript{1}, Muhammad Rafsan Kabir\textsuperscript{1}, Robin Krambroeckers\textsuperscript{1}, Sifat Momen\textsuperscript{2}, \\Nabeel Mohammed\textsuperscript{2}, Shafin Rahman\textsuperscript{2}}

\IEEEauthorblockA{\textsuperscript{1}Artificial Intelligence Department, RobotBulls Labs, Geneva, Switzerland}
\IEEEauthorblockA{\textsuperscript{2}Machine Intelligence Lab (MILab), North South University, Dhaka, Bangladesh}
\textsuperscript{1}\{jawad, muhammad, robin\}@robotbulls.com \\
\textsuperscript{2}\{sifat.momen, nabeel.mohammed, shafin.rahman\}@northsouth.edu
}

\begin{document}

\maketitle

\thispagestyle{plain}
\pagestyle{plain}

\begin{tikzpicture}[remember picture,overlay]
\node at ([yshift=-1cm]current page.north) {
\small
 2025 IEEE International Conference on Big Data (BigData)};
\end{tikzpicture}

\begin{abstract}
Natural Language Processing (NLP) has transformed the financial industry, enabling advancements in areas such as textual analysis, risk management, and forecasting. Large language models (LLMs) like BloombergGPT and FinMA have set new benchmarks across various financial NLP tasks, including sentiment analysis, stock movement prediction, and credit risk assessment. Furthermore, FinMA-ES, a bilingual financial LLM, has also demonstrated strong performance using the FLARE and FLARE-ES benchmarks. However, the high computational demands of these models limit the accessibility of many organizations. To address this, we propose Layer-wise Adaptive Ensemble Tuning (LAET), a novel strategy that selectively fine-tunes the most effective layers of pre-trained LLMs by analyzing hidden state representations while freezing less critical layers. LAET significantly reduces computational overhead while enhancing task-specific performance. Our approach shows strong results in financial NLP tasks, outperforming existing benchmarks and state-of-the-art LLMs such as GPT-4, even with smaller LLMs ($\sim$3B parameters). This work bridges cutting-edge financial NLP research and real-world deployment with efficient and scalable models for financial applications. Code is available at: \url{https://github.com/EncryptedBinary/LAET}.
\end{abstract}


\section{Introduction}

\noindent Natural Language Processing (NLP) has become a transformative tool across various domains, enabling advancements in tasks such as sentiment analysis, named entity recognition (NER), prediction, and question-answering (QA). With the support of machine learning, it is now possible to extract valuable insights from unstructured text sources like reports, news articles, and social media. However, domain-specific texts often contain technical language, specialized terminology, and complex structures that challenge the performance of general-purpose language models~\cite{araci2019finbert, xie2023wall, han2023select}. This has led to growing interest in adapting NLP techniques to domain-specific contexts for improved accuracy and reliability.

\begin{figure}[!t]
    \centering
    \includegraphics[width=0.5\textwidth]{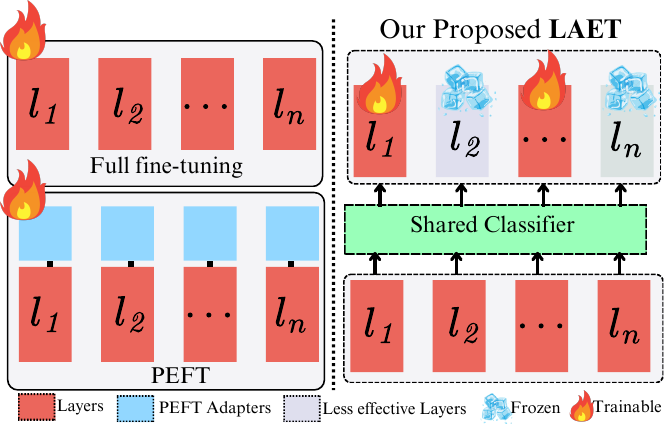}
    \caption{Comparison of full fine-tuning, parameter-efficient fine-tuning (PEFT), and our proposed Layer-wise Adaptive Ensemble Tuning (LAET). LAET identifies and updates only the most effective layers while keeping all remaining layers frozen. This substantially reduces computational overhead compared to full fine-tuning. At the same time, it achieves higher performance than conventional PEFT approaches (eg, LoRA, DoRA, etc), which tune only small adapter modules and later merge their weights.}
    \label{fig_1}
\end{figure}

\noindent Recent progress in Large Language Models (LLMs), such as GPT-4, has demonstrated strong generalization capabilities across domains, including healthcare, law, programming, and finance~\cite{lopez2023can,brown2020language, bubeck2023sparks}. These models can perform complex tasks by following natural language instructions with minimal fine-tuning. Yet, in highly specialized domains, their effectiveness is often limited by a lack of domain-specific understanding and evaluation benchmarks~\cite{xie2023wall}. To address this, researchers have introduced domain-adapted models like finBERT~\cite{araci2019finbert}, FinBERT~\cite{yang2020finbert}, and FLANG~\cite{shah2022flue}, which are pre-trained on targeted datasets. While useful, these models typically have smaller parameter sizes (under one billion), limiting their adaptability to broader tasks. More recently, large-scale models such as BloombergGPT~\cite{wu2023bloomberggpt} have emerged, showing the potential of domain-specific LLMs trained at scale.

\noindent Despite these developments, practical deployment of domain-specific language models still faces key challenges:
\textbf{(a) Architectural complexity} – advanced models like BloombergGPT~\cite{wu2023bloomberggpt}, FinGPT~\cite{yang2023fingpt}, PIXIU~\cite{xie2023pixiu}, FinBen~\cite{xie2024finben}, and multilingual FinMA-ES~\cite{zhang2024dolares} deliver strong results but require highly complex and resource-intensive architectures.
\textbf{(b) High resource demands }– training and fine-tuning these large models is computationally expensive and often inaccessible to smaller organizations.
\textbf{(c) Inefficient layer usage} – emerging evidence suggests that many layers in large LLMs contribute little to performance, leading to unnecessary computation~\cite{jin2024exploring, gromov2403unreasonable, fan2024not}.

\noindent Recent innovations in compact models like Phi~\cite{abdin2024phi}, Gemma~\cite{team2024gemma}, and LLaMA~\cite{dubey2024llama} offer promising alternatives. These lightweight models reduce computational overhead while retaining strong performance, making them more accessible and easier to deploy. To address the efficiency-performance trade-off, this paper proposes a novel methodology for optimizing pre-trained language models by leveraging \textbf{layer-wise representations}. By identifying the most impactful layers for a given task, we fine-tune only those layers while freezing the rest, reducing computational costs. An intermediate neural network further aggregates selected features to enhance performance. Fig.~\ref{fig_1} depicts our proposed method \textbf{LAET}. Its selective adaptation strategy enables more efficient and scalable use of language models across specialized tasks.

The key contributions are as follows: 
\begin{itemize}
    \item  Proposing LAET: a layer-wise adaptation technique to maximize utility from individual layers.
    \item Incorporating an ensemble-based fine-tuning strategy to enhance overall effectiveness.
    \item Achieving strong performance on multi-domain tasks using smaller LLMs .
\end{itemize}

\section{Related Works}

\noindent\textbf{Depth Knowledge and Concepts in LLMs:} The strong performance of LLMs has sparked debate over whether they truly understand concepts or simply mimic patterns. Studies show that LLMs encode structured knowledge internally~\cite{soun2022accurate, azaria2023internal}, and altering specific components can impact reasoning~\cite{geva2023dissecting}, indicating concept-specific structures. Layer-wise analyses reveal that not all layers are equally important~\cite{zhao2024opening, fan2024not}, with pruning~\cite{gromov2403unreasonable, men2024shortgpt} and probing~\cite{alain2016understanding} methods exposing redundancy. Building on this, we propose a concept depth analysis approach that selectively trains the most relevant layers, enabling compact LLMs to perform competitively with lower computational cost.

\noindent\textbf{Parameter-Efficient Fine-tuning: }Parameter-Efficient Fine-Tuning (PEFT) methods aim to reduce the high cost of adapting large language models by updating only a small subset of parameters. These include \textbf{adapter-based} approaches~\cite{houlsby2019parameter} that insert lightweight trainable modules into frozen backbones, and \textbf{prompt-based} methods that optimize soft tokens prepended to inputs. Among PEFT techniques, \textbf{LoRA}~\cite{hu2022lora} introduces trainable low-rank matrices into attention layers, \textbf{DoRA}~\cite{liu2024dora} decouples weight norm and direction to fine-tune only directional components, and \textbf{AdaLoRA}~\cite{zhang2023adalora} adapts rank dynamically during training. However, these methods typically tune all layers uniformly, overlooking the layer-wise relevance and introducing redundancy. To overcome these limitations, we propose \textbf{LAET}, which leverages concept-depth analysis to selectively fine-tune only the most semantically critical layers. This targeted adaptation reduces cost while maintaining strong performance.

\section{Methodology}
\begin{figure*}[!t]
    \centering
    \includegraphics[width=1\textwidth]{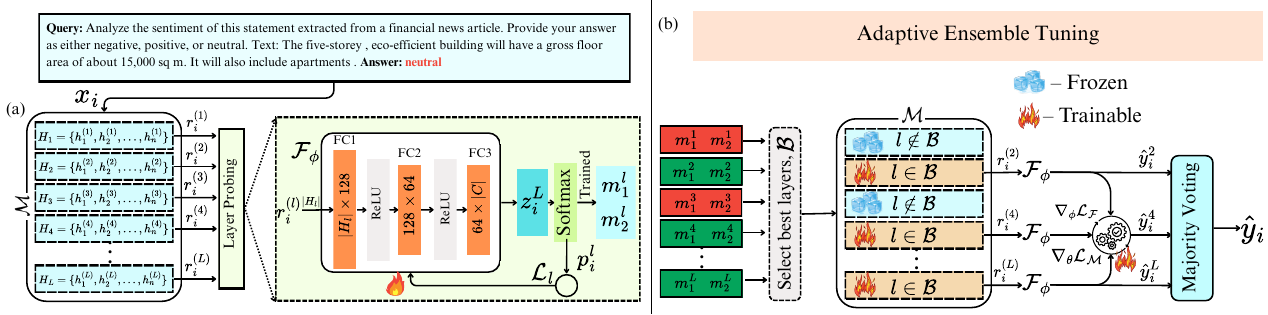}
    \caption{\textit{\textbf{LAET}}: Initially, (a) for each layer \(l\) in a pre-trained model \( \mathcal{M} \), representations \( \mathbf{r}_i^{(l)} \) are extracted, logits \( z_i^l \) are computed using the classifier \( \mathcal{F}_\phi \), and a cross-entropy loss \( \mathcal{L}_l \) is calculated. Next, (b) the best-performing layers are selected based on evaluation metrics and their deviations from the maximum values. Layers \( l \notin\mathcal{B} \) are frozen, while the selected layers and shared classifier are fine-tuned. During inference, a voting-based ensemble strategy aggregates predictions from the selected layers to generate the final output.}
    \label{fig_2}
\end{figure*}

\subsection{Problem Formulation}
\noindent \textbf{Hidden state representations: }Let \( \mathcal{M} \) be a pre-trained large language model (LLM) with \( L \) layers. The corresponding tokenizer \( \mathcal{T} \) maps an input text sequence \( \mathbf{x} \) to token IDs, such that:
\(
\mathcal{T}(\mathbf{x}) = \{\mathbf{t}_1, \mathbf{t}_2, \dots, \mathbf{t}_n\},
\)
where \( n \) represents the number of tokens resulting from the tokenization process. For each layer \( l \in \{1, 2, \dots, L\} \), the model \( \mathcal{M} \) computes hidden state representations for each token. The hidden states at layer \( l \) for all tokens are denoted as:
\( \mathbf{H}_l = \{\mathbf{h}_1^{(l)}, \mathbf{h}_2^{(l)}, \dots, \mathbf{h}_n^{(l)}\} \)
where \( \mathbf{h}_i^{(l)} \in \mathbb{R}^{d} \) represents the hidden state at layer \( l \) for token \( j \), with \( d \) being the hidden state dimension. For each layer \( l \), the representation of the last token is defined as:
\(
\mathbf{r}_l = \mathbf{r}_n^{(l)} = \mathbf{h}_n^{(l)}[:, r], \quad \mathbf{r}_l \in \mathbb{R}^d.
\)

\noindent \textbf{Implication: }Consider a dataset \( \mathcal{D} = \{(x_i, y_i)\}_{i=1}^{N} \), where each \( x_i \in \mathbb{R}^d \) is an input text and \( y_i \in \mathcal{C} \) is the corresponding label, with \( \mathcal{C} = \{c_1, c_2, \dots, c_k\} \) representing the set of \( k \) possible classes. For each input \( x_i \), the last token representation \( \mathbf{r}_i^{(l)}. \) from layer \( l \) is computed. A small neural network classifier, denoted \( \mathcal{F}_\phi \), parameterized by \( \phi \), is employed to classify the last token representations from all layers using a single set of shared weights. The logits for each input representation \( \mathbf{r}_i^{(l)} \) from layer \( l \) are computed as:
\(
z_i^l = \mathcal{F}_\phi(\mathbf{r}_i^{(l)}), \quad z_i^l \in \mathbb{R}^k.
\) Using the predicted class probabilities \( p_i^l[j] \) for each class \( j \in \mathcal{C} \), the neural network is trained to evaluate the performance of each layer. Based on these performance evaluations, the best-performing layers, denoted as \( \mathcal{B} \), are identified. Rather than training the entire model \( \mathcal{M} \), the approach focuses on leveraging the performance of the best layers \( \mathcal{B} \) for downstream tasks. To achieve this, we freeze the weights of the non-selected layers and train only the best layers \( \mathcal{B} \), alongside the shared classifier \( \mathcal{C}_\phi \). The final prediction \( \hat{y} \) is computed as the majority vote over the outputs from the best-performing layers \( \mathcal{B} \).


\subsection{Layer-wise Adaptive Ensemble Tuning (LAET)}
\noindent We propose \textbf{LAET}, a selective fine-tuning strategy that identifies and optimizes the most effective layers while incorporating ensemble decision-making, illustrated in Fig.~\ref{fig_2}. The objective of LAET is to enhance the task-specific adaptation of smaller models with minimal computational overhead.

\noindent \textbf{Hidden states: }
The model \( \mathcal{M} \) converts each token \( t_i \in \mathcal{T}(x) \) into a dense vector representation using an embedding matrix \( E \in \mathbb{R}^{|V| \times d} \), where \( |V| \) is the size of the vocabulary and \( d \) is the embedding dimension. For each token \( t_i \), the embedding \( \mathbf{e}_i \in \mathbb{R}^d \) is:
\(
\mathbf{e}_i = E[t_i].
\) Thus, the entire token sequence is transformed into an embedding matrix,
\(
\mathbf{E} = \{\mathbf{e}_1, \mathbf{e}_2, \dots, \mathbf{e}_n\}, \quad \mathbf{E} \in \mathbb{R}^{n \times d}.
\) The representations of the input tokens are initialized with these token embeddings \( \mathbf{E} \), which incorporate information about the identity of each token. A positional encoding matrix \( \mathbf{P} \in \mathbb{R}^{n \times d} \) is added to the token embeddings to inject information about the position of each token in the sequence, resulting in the following input representations:
\(
\mathbf{r}^{(0)} = \mathbf{E} + \mathbf{P},
\) where \( \mathbf{r}^{(0)} \in \mathbb{R}^{n \times d} \) is the matrix of token representations that serve as the input to the first transformer layer. These representations \( \mathbf{r}^{(0)} \) are passed through multiple transformer layers \( {L} \). For each layer \( l \in {L} \), the model computes new representations for each token based on self-attention and feedforward transformations. Specifically, for each token \( i \), the representation at layer \( l \), denoted as \( \mathbf{r}_i^{(l)} \), is derived as follows. In the self-attention mechanism, for each token \( i \), the model computes three vectors: \textit{(a)} Query vector: \( \mathbf{q}_i = W_Q \mathbf{h}_i^{(l-1)} \), \textit{(b)} Key vector: \( \mathbf{k}_i = W_K \mathbf{h}_i^{(l-1)} \),\textit{(c)} Value vector: \( \mathbf{v}_i = W_V \mathbf{h}_i^{(l-1)} \), where \( \mathbf{h}_i^{(l-1)} \) is the hidden state from the previous layer for token \( i \), and \( W_Q, W_K, W_V \in \mathbb{R}^{d \times d} \) are learned projection matrices. The attention scores between token \( i \) and all other tokens \( j \) are computed.
The attention scores are normalized using softmax and applied to the value vectors \( \mathbf{v}_j \):
\(
\mathbf{h}_i^{(l)} = \sum_{j=1}^{n} \text{softmax}( \text{Attention}(i, j)) \mathbf{v}_j.
\)
This produces the updated hidden state \( \mathbf{h}_i^{(l)} \in \mathbb{R}^d \) for token \( i \) at layer \( l \). After applying the self-attention mechanism and feedforward layers, the hidden states are used to compute the new representations at each layer, denoted as \( \mathbf{r}_i^{(l)} \). Thus, the representation \( \mathbf{r}_i^{(l)} \) for token \( i \) at layer \( l \) is ultimately a transformation of the hidden state \( \mathbf{h}_i^{(l)} \).\\
\noindent\textbf{Layer Probing: }
For each input text \( x_i \in \mathbb{R}^d \) with its corresponding label \( y_i \in \mathcal{C} \), hidden representations are extracted from each layer \( l \) while keeping the model \( \mathcal{M} \) in inference mode. The final hidden representations \( \mathbf{r}_i^{(l)} \) are stored for every input, forming the set:
\(
\mathcal{R} = \left\{ \mathcal{R}_l \right\}_{l=1}^{L}, \quad \text{where} \quad \mathcal{R}_l = \left\{ (\mathbf{r}_i^{(l)}, y_i) \right\}_{i=1}^{N}.
\)
Here, \( \mathcal{R}_l \) represents the collection of hidden representations from layer \( l \), paired with their corresponding labels across all \( N \) inputs. These representations serve as input to a small neural classifier \( \mathcal{F}_\phi \), which probes each layer’s ability to encode task-relevant information.   For each representation \( \mathbf{r}_i^{(l)} \), the classifier produces logits \( z_i^l \), which are transformed into class probabilities \( p_i^l[j] \) for class \( j \in \mathcal{C} \) using the softmax function: 
\[
p_i^l[j] = \frac{\exp(z_i^l[j])}{\sum_{j'=1}^{k} \exp(z_i^l[j'])}, \quad p_i^l \in \mathbb{R}^k. 
\]
The loss for each input \( i \) at layer \( l \) is computed using cross-entropy between the predicted probabilities and the true label \( y_i \). The total loss for layer \( l \) across all inputs is given by: 
\[
\mathcal{L}_l = -\frac{1}{N} \sum_{i=1}^{N} \sum_{j=1}^{k} \mathcal{P}[y_i = j] \log(p_i^l[j])
\] where \( \mathcal{P}(y_i = j) \) is an indicator function that equals 1 if \( y_i = j \) and 0 otherwise. 

\noindent \textbf{Best Layers Selection: }
The shared classifier $\mathcal{F}_\phi$ is trained on each layer's representation. Gradient updates for the classifier \( \mathcal{F}_\phi \) are as follows:
\[
\nabla_\phi \mathcal{L}_l = \frac{1}{N} \sum_{i=1}^{N} \sum_{j=1}^{k} (p_i^l[j] - \mathcal{P}(y_i = j)) \nabla_\phi \mathcal{F}_\phi(\mathbf{r}_i^{(l)}).
\] After training, two performance metrics, based on literature, $m_1$ (e.g., accuracy) and $m_2$ (e.g., F1 score or MCC), are computed for each layer $l \in \{1, \dots, L\}$. The standard deviations $\sigma_{m_1}$ and $\sigma_{m_2}$ across all layers are computed as:
\[
\sigma_{m_1} = \text{std}(\{m_1^1, \dots, m_1^L\}), \quad \sigma_{m_2} = \text{std}(\{m_2^1, \dots, m_2^L\}).
\]. Dynamic margins are then defined as:
\(
\delta_{m_1} = \alpha \cdot \sigma_{m_1}, \quad \delta_{m_2} = \beta \cdot \sigma_{m_2}.
\) A layer $l$ is included in the final set of best-performing layers $\mathcal{B}$ if no other layer $l' \ne l$ satisfies:
\[
m_1^{l'} \ge m_1^l + \delta_{m_1} \quad \text{and} \quad m_2^{l'} \ge m_2^l + \delta_{m_2}.
\]. The resulting subset $\mathcal{B}$ defines the layers used during fine-tuning. To formally analyze the optimization behavior under this selection, consider the modified training objective:
\[
\mathcal{L}_{\mathcal{B}}(\theta_{\mathcal{B}}, \phi) = \frac{1}{|\mathcal{B}|} \sum_{l \in \mathcal{B}} \mathcal{L}^{(l)}(\theta_l, \phi),
\], where $\theta_{\mathcal{B}} = \{\theta_l\}_{l \in \mathcal{B}}$ denotes the parameters of the selected layers, and $\phi$ represents the shared classifier parameters. Each loss term $\mathcal{L}^{(l)}$ corresponds to the output computed from the representation $\mathbf{r}^{(l)}$ of layer $l$. Assuming that $\mathcal{L}^{(l)}$ is differentiable with Lipschitz-continuous gradients and that updates are performed using stochastic gradient descent with a diminishing learning rate $\eta_t$ satisfying $\sum_t \eta_t = \infty$ and $\sum_t \eta_t^2 < \infty$, it follows from standard stochastic approximation theory that:
\[
\lim_{t \to \infty} \mathbb{E}\left[\left\| \nabla_{\theta_{\mathcal{B}}} \mathcal{L}_{\mathcal{B}} \right\|^2 + \left\| \nabla_{\phi} \mathcal{L}_{\mathcal{B}} \right\|^2\right] = 0.
\]
This establishes that convergence to a stationary point is guaranteed despite partial updates to a subset of layers. In addition to its convergence properties, the proposed standard deviation-based selection method offers greater robustness than gradient-based alternatives. Approaches that rely on gradient norm magnitudes, such as $\|\nabla_{\theta_l} \mathcal{L}\|$, are often sensitive to depth-related variance and may not correlate with downstream performance. In contrast, LAET defines selection thresholds concerning validation performance, ensuring that only layers exhibiting statistically stable and semantically aligned behavior are selected. Specifically, the retained set of layers satisfies \(\mathcal{B}\). Empirically, this approach reduces the number of trainable layers without compromising accuracy. Theoretically, it can be interpreted as selecting from the Pareto frontier of performance metrics, in contrast to gradient-based methods that do not encode task-level validation behavior. This strategy boosts both the efficiency and generalization of LAET. 



\noindent \textbf{Fine tune strategy: }Once the best-performing layers \( \mathcal{B} \) are identified, the remaining layers, \( l \notin \mathcal{B} \), are frozen, while the selected layers, \( l \in \mathcal{B} \) within $\mathcal{M}$, along with the shared-weight classifier \( \mathcal{F}_\phi \), undergo fine-tuning. Since the classifier is shared across all selected layers, the total loss across \( \mathcal{B} \) is computed as:  
\(
\mathcal{L} = \frac{1}{|\mathcal{B}|} \sum_{l \in \mathcal{B}} \mathcal{L}_l.
\)
The gradient of the total loss with respect to the classifier parameters \( \phi \) is given by:  
\[
\nabla_\phi \mathcal{L} = \frac{1}{|\mathcal{B}|} \sum_{l \in \mathcal{B}} \nabla_\phi \mathcal{L}_l.
\]
Expanding \( \nabla_\phi \mathcal{L}_l \) using the softmax derivative:  
\[
\nabla_\phi \mathcal{L}_l = \frac{1}{N} \sum_{i=1}^{N} \sum_{j=1}^{k} (p_i^l[j] - \mathcal{P}(y_i = j)) \nabla_\phi z_i^l[j].
\]
Since the logits \( z_i^l \) are produced by the classifier, \( z_i^l = \mathcal{F}_\phi(\mathbf{r}_i^{(l)}) \), applying the chain rule yields:  
\[
\nabla_\phi z_i^l[j] = \frac{\partial \mathcal{F}_\phi(\mathbf{r}_i^{(l)})}{\partial \phi}.
\]
Thus, the final gradient update for the shared classifier \( \mathcal{F}_\phi \) is:  
\[
\nabla_\phi \mathcal{L}_\mathcal{F} = \frac{1}{|\mathcal{B}| N} \sum_{l \in \mathcal{B}} \sum_{i=1}^{N} \sum_{j=1}^{k} (p_i^l[j] - \mathcal{P}(y_i = j)) \nabla_\phi \mathcal{F}_\phi(\mathbf{r}_i^{(l)}).
\]
For the model \( \mathcal{M} \) with parameters \( \theta \), updates are applied only to the selected layers \( \mathcal{B} \). The gradient with respect to \( \theta_l \) for each \( l \in \mathcal{B} \) is computed as:  
\[
\nabla_{\theta_l} \mathcal{L}_{\mathcal{M}} = \frac{1}{|\mathcal{B}| N} \sum_{i=1}^{N} \sum_{j=1}^{k} (p_i^l[j] - \mathcal{P}(y_i = j)) \nabla_{\theta_l} \mathbf{r}_i^{(l)}.
\]
\noindent \textbf{Voting-based Ensembled for Prediction: } After training both the classifier $\mathcal{F}_\phi$ and the model $\mathcal{M}$, the final prediction is obtained through an ensemble over the selected layers $\mathcal{B}$. For a given unseen input \( \mathbf{x}_u \), each layer \( l \in \mathcal{B} \) produces a prediction \( \hat{y}_l = \arg\max_{j} p_u^l[j] \), where \( p_u^l = \text{softmax}(\mathcal{F}_\phi(\mathbf{r}_u^{(l)})) \) is the class probability vector. Instead of relying on any single layer, a majority voting rule is employed to aggregate predictions across all selected layers. The final predicted class is given:
\[
\hat{y} = \arg \max_{c \in \mathcal{C}} \sum_{l \in \mathcal{B}} \mathbb{I}(\hat{y}_l = c),
\] where \( \mathbb{I}(\cdot) \) is the indicator function. This procedure reduces sensitivity to any single layer's misprediction and exploits the diversity among informative layer representations. Under the assumption that predictions \( \hat{y}_l \) from different layers are conditionally independent, and that each layer predictor has a generalization error rate bounded by \( \epsilon_l = \mathcal{P}(\hat{y}_l \ne y) \), the expected ensemble error can be analyzed using classical results from ensemble learning. Let 
\( \bar{\epsilon} = \frac{1}{|\mathcal{B}|} \sum_{l \in \mathcal{B}} \epsilon_l \) 
denote the average error rate. If \( \bar{\epsilon} < 0.5 \), then the error of the ensemble prediction satisfies the exponential bound:\[
\mathcal{P}(\hat{y} \ne y) \leq \exp\left(-2|\mathcal{B}| (0.5 - \bar{\epsilon})^2\right)
\]. This result shows that ensemble error decreases rapidly as more accurate layers are included in the vote. The generalization error of each layer predictor is influenced by the complexity of the shared classifier \( \mathcal{F}_\phi \). Let \( \mathcal{H} \) denote the hypothesis class defined by \( \mathcal{F}_\phi \), with VC-dimension \( \mathrm{VC}(\mathcal{H}) \). Assume that each layer predictor is trained using \( N \) probing samples. Then, for any layer \( l \) and for any \( \delta > 0 \), the probability that the difference between the empirical error \( \hat{\epsilon}_l(h) \) and the true error \( \epsilon_l(h) \) exceeds \( \delta \) is bounded by:
\[ \mathcal{P}\left( \left| \hat{\epsilon}_l(h) - \epsilon_l(h) \right| > \delta \right) \leq 2 \cdot \left( \frac{1}{N} \right)^{2\delta^2 - \frac{\mathrm{VC}(\mathcal{H}) \log N}{N}}.\]
This implies that with enough probing samples and a moderately complex classifier, each layer predictor generalizes reliably, supporting the stability of the majority vote ensemble.

\begin{algorithm}[!t]
\scriptsize
\caption{LAET}
\label{alg:alg_1}
\begin{algorithmic}[1]

\STATE \textbf{Note:} A pre-trained model \( \mathcal{M} \) with \( L \) layers, dataset \( \mathcal{D} \), classifier \( \mathcal{F}_\phi \), selection thresholds \( \alpha, \beta \); \textbf{Purpose:} fine-tune classifier \( \mathcal{F}_\phi \) and selected layers \( \mathcal{B} \).
\STATE \textbf{Input:} $x_i \in \mathcal{D}$ where, $x_i \in \mathbb{R}^d$.
\STATE \textbf{Output:} Classification $\hat{y}_i \in \mathcal{C}$.
\STATE \textbf{Step 1: Train Classifier for Each Layer}
\FOR{$l = 1$ to $L$}
    \STATE Extract representation: $\mathbf{r}_i^{(l)} \gets \mathcal{M}_l(x_i)$
    \STATE Compute logits: $z_i^l \gets \mathcal{F}_\phi(\mathbf{r}_i^{(l)})$
    \STATE Compute probability: $p_i^l \gets \text{Softmax}(z_i^l)$
    \STATE Compute loss: $\mathcal{L}_l \gets \text{CrossEntropy}(p_i^l, y_i)$
    \STATE Update classifier: $\phi \gets \phi - \eta \nabla_\phi \mathcal{L}_l$
\ENDFOR

\STATE \textbf{Step 2: Select Best Layers}
\STATE Compute metrics: $m_1^l, m_2^l \quad \forall l$
\STATE Compute standard deviations: $\sigma_{m_1}, \sigma_{m_2}$
\STATE Select best layers: $\mathcal{B} \gets \{ l \mid m_1^l \geq \max m_1 - \alpha \sigma_{m_1}, m_2^l \geq \max m_2 - \beta \sigma_{m_2} \}$

\STATE \textbf{Step 3: Freeze Layers Except Best Layers}
\FOR{$l \notin \mathcal{B}$}
    \STATE Freeze layer: $\mathcal{M}_l \gets \text{Frozen}$
\ENDFOR

\STATE \textbf{Step 4: Fine-Tune Best Layers and Classifier}
\FOR{each training step}
    \FOR{$l \in \mathcal{B}$}
        \STATE Extract representation: $\mathbf{r}_i^{(l)} \gets \mathcal{M}_l(x_i)$
        \STATE Compute logits: $z_i^l \gets \mathcal{F}_\phi(\mathbf{r}_i^{(l)})$
        \STATE Compute probability: $p_i^l \gets \text{Softmax}(z_i^l)$
        \STATE Compute loss: $\mathcal{L}_l \gets \text{CrossEntropy}(p_i^l, y_i)$
    \ENDFOR
    \STATE Compute total loss: $\mathcal{L} \gets \frac{1}{|\mathcal{B}|} \sum_{l \in \mathcal{B}} \mathcal{L}_l$
    \STATE Update classifier: $\phi \gets \phi - \eta \nabla_\phi \mathcal{L}$
    \STATE Update selected layers: $\theta_{\mathcal{B}} \gets \theta_{\mathcal{B}} - \eta \nabla_{\theta_{\mathcal{B}}} \mathcal{L}$
\ENDFOR

\STATE \textbf{Step 5: Voting-Based Ensemble Prediction}
\FOR{each input $x_u$}
    \STATE Compute predictions: $\hat{y}_l \gets \mathcal{F}_\phi(\mathbf{r}_u^{(l)}) \quad \forall l \in \mathcal{B}$
    \STATE Compute final prediction: $\hat{y} \gets \arg\max_c \sum_{l \in \mathcal{B}} \mathbb{I}(\hat{y}_l = c)$
\ENDFOR

\end{algorithmic}
\end{algorithm}

\begin{table*}[!t]
\tiny
\caption{Statistics of the employed benchmark datasets along with their corresponding evaluation metrics.}
\label{tab:table_1}
\centering
\scriptsize
\begin{tabular}{lllrrrrrllllr}
\toprule
\textbf{Datasets} & \textbf{Task} & \textbf{Language}  & \textbf{Data Type} & \textbf{Modalities} & \textbf{Train} & \textbf{Val.} & \textbf{Test} & \textbf{Evaluation} \\
\midrule
FPB~\cite{malo2014good} & sentiment analysis & English & news & text & 3100 & 776 & 970 & Accuracy, F1  \\
FiQA-SA~\cite{maia201818} & sentiment analysis & English & news headlines, tweets & text & 750 & 188 & 235 & Accuracy, F1 \\
TSA~\cite{cortis2017semeval} & sentiment analysis & English & news headlines & text & 448 & -- & 113 & RMSE \\
TSA~\cite{pan2023evaluation} & sentiment analysis & Spanish & news headlines & text & 3063 & -- & 766 & Accuracy, F1 \\
FinanceES~\cite{nguyen2023abcd} & sentiment analysis & Spanish & news headlines & text & 5084 & -- & 1272 & Accuracy, F1 \\
Headlines~\cite{sinha2021impact} & news headline classification & English & news headlines & text & 16437 & -- & 4110 & Accuracy, Avg F1 \\
FOMC~\cite{shah2023trillion} & hawkish-dovish classification & English & news & text & 396 & -- & 100 & Accuracy, F1 \\
FinArg-ACC~\cite{sy2023fine} & argument unit classification & English & comments & text & 6202 & -- & 1551 & Accuracy, Micro F1 \\
MultiFin~\cite{jorgensen2023multifin} & multi-class classification & English & article headlines & text & 5350 & -- & 1340 & Accuracy, Micro F1 \\
MultiFin~\cite{jorgensen2023multifin} & multi-class classification & Spanish & article headlines & text & 210 & -- & 20 & Accuracy, Micro F1 \\
MA~\cite{yang2020generating} & deal completeness classification & English & news, tweets & text & 400 & -- & 100 & Accuracy, Micro F1 \\
\midrule
BigData22~\cite{soun2022accurate} & stock movement prediction & English & tweets, historical prices & text, time series & 4900 & 768 & 1470 & Accuracy, F1, MCC  \\
ACL18~\cite{xu2018stock} & stock movement prediction & English & tweets, historical prices & text, time series & 20800 & 2560 & 3720 & Accuracy, F1, MCC \\
CIKM18~\cite{wu2018hybrid} & stock movement prediction & English & tweets, historical prices & text, time series & 3400 & 1140 & 431 & Accuracy, F1, MCC \\
\midrule
German~\cite{hofmannstatlog} & credit scoring & English & biography & Multivariate & 700 & 100 & 200 & Accuracy, F1, MCC \\
Australian~\cite{quinlan1987statlog} & credit scoring & English & biography & Multivariate & 482 & 69 & 139 & Accuracy, F1, MCC \\
LendingClub~\cite{feng2023empowering} & credit scoring & English & loan record & Multivariate & 9420 & 1350 & 2690 & Accuracy, F1, MCC \\
ccf~\cite{feng2023empowering} & fraud detection & English & financial record & Multivariate & 7970 & 1140 & 2280 & Accuracy, F1, MCC \\
cfraud~\cite{feng2023empowering} & fraud detection & English & financial record & text & 7340 & 1050 & 2100 & Accuracy, F1, MCC \\
polish~\cite{feng2023empowering} & financial distress identification & English & financial profile & text, table & 6080 & 868 & 1740 & Accuracy, F1, MCC \\
taiwan~\cite{feng2023empowering} & financial distress identification & English & financial profile & text, table & 4770 & 681 & 1370 & Accuracy, F1, MCC \\
ProtoSegno~\cite{feng2023empowering} & financial distress identification & English & financial profile & Multivariate & 8330 & 1190 & 2380 & Accuracy, F1, MCC \\
travrealtinsurance~\cite{feng2023empowering} & claim analysis & English & financial profile & Multivariate & 8870 & 1270 & 2530 & Accuracy, F1, MCC \\
\bottomrule
\end{tabular}
\end{table*}

\subsection{Primary Area of Assessment: Finance}
\noindent To evaluate our method, LAET, we begin with the financial domain due to its rich set of benchmark datasets for text classification and the availability of recent strong baselines. We assess across three areas: Textual Analysis (TA), Forecasting (FO), and Risk Management (RM)—each capturing distinct reasoning challenges. Together, they provide a comprehensive testbed for evaluating LLM performance. See Table~\ref{tab:table_1} for details. \textit{(1) TA} covers eight tasks including sentiment classification (using FPB~\cite{malo2014good}, FiQA-SA~\cite{maia201818}, TSA~\cite{cortis2017semeval}), headline-based price movement prediction~\cite{sinha2021impact}, policy tone classification~\cite{shah2023trillion}, argument mining~\cite{sy2023fine}, multi-class document tagging~\cite{jorgensen2023multifin}, and MA event prediction~\cite{yang2020generating}. \textit{(2) FO} involves stock movement prediction using datasets like BigData22~\cite{soun2022accurate}, ACL18~\cite{xu2018stock}, and CIKM18~\cite{wu2018hybrid}, combining financial time series with textual sentiment. \textit{(3) RM} tasks include credit scoring~\cite{hofmannstatlog,quinlan1987statlog,feng2023empowering}, fraud detection, bankruptcy prediction, and insurance claim analysis~\cite{feng2023empowering}. These domains collectively enable a comprehensive evaluation of LLM performance under real-world financial reasoning challenges.

\begin{table*}[!t]
\small
\centering
\caption{Performance comparison of LAET models based on Accuracy (Acc), F1, Average F1 (AvF1), MicroF1 (MiF1) against benchmarks on TA tasks. with the best values highlighted in bold. Unreported values are marked as (-). Highest performance is marked by \textbf{\textcolor{red}{Bold red}} and the second highest given by \textbf{\textcolor{blue}{bold blue}}. }
\label{tab: table_2}.
\setlength{\tabcolsep}{1.7pt}
\begin{adjustbox}{max width=\textwidth}
\begin{tabular}{l c c c c c c c c c c c c c c c c c c c c}
\hline
\textbf{Models} & \multicolumn{2}{c}{\textbf{FPB}} & \multicolumn{2}{c}{\textbf{FiQA.}} & \multicolumn{2}{c}{\textbf{TSA}} & \multicolumn{1}{c}{\textbf{TSA}} & \multicolumn{2}{c}{\textbf{Fin.ES}} & \multicolumn{2}{c}{\textbf{Head.}} & \multicolumn{2}{c}{\textbf{FOMC}} & \multicolumn{2}{c}{\textbf{FinArg.}} & \multicolumn{2}{c}{\textbf{MultiF}} & \multicolumn{1}{c}{\textbf{MultiF}} & \multicolumn{2}{c}{\textbf{MA}}\\ 
 \cmidrule(lr){2-3} \cmidrule(lr){4-5} \cmidrule(lr){6-7} \cmidrule(lr){8-8} \cmidrule(lr){9-10} \cmidrule(lr){11-12} \cmidrule(lr){13-14} \cmidrule(lr){15-16} \cmidrule(lr){17-18} \cmidrule(lr){19-19} \cmidrule(lr){20-21} 
 & \textbf{Acc} & \textbf{F1} & \textbf{Acc} & \textbf{F1} & \textbf{Acc} & \textbf{F1} & \textbf{RMSE} & \textbf{Acc} & \textbf{F1} & \textbf{Acc} & \textbf{AvF1} & \textbf{Acc} & \textbf{F1} & \textbf{Acc} & \textbf{MiF1} & \textbf{Acc} & \textbf{F1} & \textbf{MiF1} & \textbf{Acc} & \textbf{MiF1}\\
 \hline

\multicolumn{21}{c}{\textit{General Purpose LLM}} \\
ChatGPT~\cite{zhang2024dolares} & 0.78 & 0.78 & - & 0.6 & 0.21 & 0.24 & - & 0.13 & 0.08 & - & 0.77 & 0.6 & 0.64 & - & - & 0.48 & 0.47 & - & - & -\\ 
GPT-4~\cite{zhang2024dolares} & 0.76 & 0.78 & - & 0.8 & 0.47 & 0.56 & - & 0.15 & 0.09 & - & 0.86 & 0.69 & 0.71 & - & - & 0.6 & 0.6 & - & - & -\\ 
Gemini~\cite{xie2024finben} & 0.77 & 0.77 & - & 0.81 & - & - & 0.37 & - & - & - & 0.78 & 0.60 & 0.40 & - & 0.31& - & 0.62 & 0.62 & - & 0.84\\
LLaMA2-7B~\cite{zhang2024dolares} & 0.68 & 0.65 & - & 0.77 & 0.07 & 0.04 & - & 0.14 & 0.13 & - & 0.72 & 0.50 & 0.35 & - & 0.46  & 0.23 & 0.11 & - & - & 0.70\\ 
LLaMA2-70B~\cite{xie2024finben} & 0.73 & 0.72 & - & 0.83 & - & - & 0.57 & - & - & - & 0.63 & 0.47 & 0.49 & - & 0.58 & - & 0.63 & 0.63 & - & 0.86\\
LLaMA3-8B~\cite{xie2024finben} & 0.52 & 0.52 & - & 0.70 & - & - & 0.25 & - & - & - & 0.60 & 0.41 & 0.40 & - & 0.51  & - & 0.39 & 0.39 & - & 0.34\\
\hline
\multicolumn{21}{c}{\textit{Few-shot LLM}} \\
GPT-4o~\cite{achiam2023gpt} & 0.87 & 0.88 & 0.70 & 0.87 & 0.72 & 0.66 & 0.55 & 0.70 & 0.58 & 0.65 & 0.97 & 0.66 & 0.68 & 0.65 & 0.72 & 0.65 & 0.70 & 0.69 & 0.61 & 0.56  \\

Deepseek V3~\cite{liu2024deepseek} & 0.83 & 0.83 & 0.70 & 0.85 & 0.81 & 0.81 & 0.57 & 0.11 & 0.11 & 0.65 & 0.96 & 0.55 & 0.49 & 0.59 & 0.66 & 0.99 & 0.99 & 0.58 & 0.72 & 0.65 \\

Qwen2.5 72B~\cite{hui2024qwen2} & 0.73 & 0.73 & 0.59 & 0.83 & 0.76 & 0.78 & 0.58 & 0.13 & 0.13 & 0.67 & 0.58 & 0.50 & 0.46 & 0.72 & 0.60 & 0.91 & 0.91 & 0.72 & 0.60 & 0.55  \\
\hline
\multicolumn{21}{c}{\textit{Financial LLM}} \\
FinMA-7B~\cite{xie2024finben} & 0.88 & 0.88 & - & 0.84 & - & - & - & - & - & - & 0.98 & - & - & - & - & - & - & - & - & -\\
FinMA-7B~\cite{xie2023pixiu} & 0.86 & 0.86 & - & 0.79 & - & - & 0.80 & - & - & - & 0.97 & 0.47 & 0.49 & - & 0.27 & - & 0.14 & 0.14 & - & 0.45\\
FinGPT-7Bl~\cite{xie2024finben} & 0.00 & 0.00 & - & 0.00 & - & - & 0.00 & - & - & - & 0.60 & 0.00 & 0.00 & - & 0.00  & - & 0.00 & 0.00 & - & 0.00\\
FinMA-30B~\cite{xie2023pixiu} & 0.87 & 0.88 & - & 0.87 & - & - & - & - & - & - & 0.97 & - & - & - & - & - & - & - & - & -\\
FinMA-ESB~\cite{zhang2024dolares} & 0.83 & 0.83 & - & 0.85 & 0.85 & 0.86 & - & 0.11 & 0.11 & - & 0.96 & 0.55 & 0.49 & - & - & 0.99 & 0.99 & - & - & - \\ 
FinMA-ESS~\cite{zhang2024dolares} & 0.73 & 0.73 & - & 0.83 & 0.86 & 0.86 & - & 0.13 & 0.13 & - & 0.58 & 0.50 & 0.46 & - & - & 0.91 & 0.91 & - & - & - \\ 
\hline
\multicolumn{21}{c}{\textit{PEFT Baselines}} \\

$\mathcal{M}_1$ - LoRA~\cite{hu2022lora} & 0.84 & 0.83 & 0.84 & 0.83 & 0.80 & 0.79 & 0.20 & 0.75 & 0.74 & 0.92 & 0.92 & 0.66 & 0.65 & 0.71 & 0.70 & 0.86 & 0.86 & 0.83 & 0.82 & 0.82 \\ 
$\mathcal{M}_2$ - LoRA~\cite{hu2022lora} & 0.85 & 0.85 & 0.84 & 0.84 & 0.80 & 0.79 & 0.22 & 0.70 & 0.70 & 0.92 & 0.92 & 0.66 & 0.66 & 0.72 & 0.72 & 0.86 & 0.86 & 0.88 & 0.84 & 0.84 \\  
$\mathcal{M}_3$ - LoRA~\cite{hu2022lora} & 0.80 & 0.81 & 0.86 & 0.86 & 0.80 & 0.79 & 0.17 & 0.72 & 0.71 & 0.93 & 0.93 & 0.66 & 0.66 & 0.68 & 0.68 & 0.84 & 0.83 & 0.89 & 0.85 & 0.85 \\
$\mathcal{M}_1$ - DoRA~\cite{liu2024dora} & 0.85 & 0.84 & 0.85 & 0.84 & 0.81 & 0.81 & 0.20 & 0.77 & 0.76 & 0.94 & 0.94 & 0.66 & 0.66 & 0.73 & 0.72 & 0.87 & 0.87 & 0.84 & 0.83 & 0.83 \\ 
$\mathcal{M}_2$ - DoRA~\cite{liu2024dora} & 0.86 & 0.86 & 0.85 & 0.85 & 0.81 & 0.81 & 0.22 & 0.72 & 0.72 & 0.94 & 0.94 & 0.68 & 0.68 & 0.74 & 0.74 & 0.87 & 0.87 & 0.90 & 0.85 & 0.85 \\  
$\mathcal{M}_3$ - DoRA~\cite{liu2024dora} & 0.86 & 0.86 & 0.86 & 0.86 & 0.81 & 0.81 & 0.17 & 0.74 & 0.73 & 0.94 & 0.94 & 0.68 & 0.68 & 0.70 & 0.70 & 0.85 & 0.84 & 0.91 & 0.87 & 0.87 \\ 
$\mathcal{M}_1$ - AdaLoRA~\cite{zhang2023adalora} & 0.81 & 0.80 & 0.81 & 0.80 & 0.77 & 0.76 & 0.23 & 0.70 & 0.68 & 0.90 & 0.90 & 0.63 & 0.62 & 0.68 & 0.66 & 0.82 & 0.82 & 0.79 & 0.78 & 0.78 \\ 
$\mathcal{M}_2$ - AdaLoRA~\cite{zhang2023adalora} & 0.82 & 0.82 & 0.82 & 0.82 & 0.77 & 0.76 & 0.25 & 0.67 & 0.67 & 0.91 & 0.91 & 0.64 & 0.63 & 0.70 & 0.70 & 0.83 & 0.82 & 0.84 & 0.81 & 0.81 \\ 
$\mathcal{M}_3$ - AdaLoRA~\cite{zhang2023adalora} & 0.81 & 0.82 & 0.83 & 0.83 & 0.77 & 0.76 & 0.24 & 0.69 & 0.69 & 0.91 & 0.91 & 0.65 & 0.64 & 0.66 & 0.66 & 0.81 & 0.80 & 0.85 & 0.82 & 0.82 \\ 
\hline
\multicolumn{21}{c}{\textit{Our Approach (LAET)}} \\
$\mathcal{M}_1$ - LAET (Ours) & 0.88 & 0.87 & 0.88 & 0.87 & 0.84 & 0.83 & \textcolor{blue}{\textbf{0.21}} & \textcolor{blue}{\textbf{0.79}} & \textcolor{blue}{\textbf{0.78}} & 0.97 & 0.97 & 0.70 & 0.68 & \textcolor{red}{\textbf{0.75}} & \textcolor{red}{\textbf{0.74}} & \textcolor{red}{\textbf{0.90}} & \textcolor{red}{\textbf{0.90}} & 0.87 & 0.86 & 0.86 \\

$\mathcal{M}_2$ - LAET (Ours) & \textcolor{red}{\textbf{0.89}} & \textcolor{red}{\textbf{0.89}} & \textcolor{red}{\textbf{0.88}} & \textcolor{red}{\textbf{0.88}} & \textcolor{red}{\textbf{0.84}} & \textcolor{red}{\textbf{0.83}} & 0.23 & 0.74 & 0.74 & \textcolor{red}{\textbf{0.97}} & \textcolor{red}{\textbf{0.97}} & \textcolor{red}{\textbf{0.70}} & \textcolor{blue}{\textbf{0.70}} & \textcolor{blue}{\textbf{0.76}} & \textcolor{blue}{\textbf{0.76}} & \textcolor{blue}{\textbf{0.90}} & \textcolor{blue}{\textbf{0.90}} & \textcolor{red}{\textbf{0.93}} & \textcolor{red}{\textbf{0.88}} & \textcolor{red}{\textbf{0.88}} \\

$\mathcal{M}_3$ - LAET (Ours) & \textcolor{red}{\textbf{0.89}} & \textcolor{blue}{\textbf{0.89}} & \textcolor{blue}{\textbf{0.90}} & \textcolor{blue}{\textbf{0.90}} & \textcolor{blue}{\textbf{0.84}} & \textcolor{blue}{\textbf{0.83}} & \textcolor{red}{\textbf{0.18}} & \textcolor{red}{\textbf{0.76}} & \textcolor{red}{\textbf{0.75}} & \textcolor{blue}{\textbf{0.98}} & \textcolor{blue}{\textbf{0.98}} & \textcolor{blue}{\textbf{0.70}} & \textcolor{red}{\textbf{0.69}} & 0.72 & 0.72 & 0.88 & 0.87 & \textcolor{blue}{\textbf{0.94}} & \textcolor{blue}{\textbf{0.89}} & \textcolor{blue}{\textbf{0.89}} \\
\hline

\end{tabular}
\end{adjustbox}
\end{table*}

\begin{table*}[!t]
\centering
\caption{Performance comparison of LAET models based on Accuracy(Acc), F1, and MCC, against benchmarks on RM tasks. with the best values highlighted in bold. Unreported values are marked as (-). Highest performance is marked by \textbf{\textcolor{red}{Bold red}} and the second highest given by \textbf{\textcolor{blue}{bold blue}}. }
\label{tab: table_3}
\setlength{\tabcolsep}{1.2pt}
\begin{adjustbox}{max width=\textwidth}
\begin{tabular}{l c c c c c c c c c c c c c c c c c c c c c c c c c c c}
\toprule
\textbf{Models} & \multicolumn{3}{c}{\textbf{German}} & \multicolumn{3}{c}{\textbf{Australian}} &
\multicolumn{3}{c}{\textbf{LendingClub}} & \multicolumn{3}{c}{\textbf{ccf}} & \multicolumn{3}{c}{\textbf{ccfraud}} & \multicolumn{3}{c}{\textbf{polish}} & \multicolumn{3}{c}{\textbf{taiwan}} & \multicolumn{3}{c}{\textbf{ProtoSegno}} & \multicolumn{3}{c}{\textbf{travr.}}\\ 
\cmidrule(lr){2-4} \cmidrule(lr){5-7} \cmidrule(lr){8-10} \cmidrule(lr){11-13} \cmidrule(lr){14-16} \cmidrule(lr){17-19} \cmidrule(lr){20-22} \cmidrule(lr){23-25} \cmidrule(lr){26-28}
 & \textbf{Acc} & \textbf{F1} & \textbf{MCC} & \textbf{Acc} & \textbf{F1} & \textbf{MCC} & \textbf{Acc} & \textbf{F1} & \textbf{MCC} & \textbf{Acc} & \textbf{F1} & \textbf{MCC} & \textbf{Acc} & \textbf{F1} & \textbf{MCC} & \textbf{Acc} & \textbf{F1} & \textbf{MCC} & \textbf{Acc} & \textbf{F1} & \textbf{MCC} & \textbf{Acc} & \textbf{F1} & \textbf{MCC} & \textbf{Acc} & \textbf{F1} & \textbf{MCC}\\
 \midrule

\multicolumn{28}{c}{\textit{General purpose LLM}} \\
ChatGPT~\cite{zhang2024dolares} & 0.2 & 0.41 & - & 0.41 & 0.26 & - & - & - & - & - & - & - & - & - & - & - & - & - & - & - & - & - & - & - & - & - & -\\ 
GPT-4~\cite{zhang2024dolares} & 0.55 & 0.513 & - & 0.74 & 0.75 & - & - & - & - & - & - & - & - & - & - & - & - & - & - & - & - & - & - & - & - & - & -\\ 
Gemini~\cite{xie2024finben} & - & 0.52 & 0.00 & - & 0.26 & 0.00 & - & 0.65 & 0.19 & - & 0.96 & -0.01 & - & 0.90 & 0.00 & - & 0.86 & \textbf{0.14} & - & 0.95 & 0.00 & - & 0.95 & 0.00 & - & 0.00 & 0.00\\
LLaMA2-7B~\cite{zhang2024dolares} & 0.61 & 0.60 & - & - & -  & - & - & -  & - & - & -  & - & - & -  & - & - & - & - & - & - & - & - & - & - & - & - & -\\ 
LLaMA2-70B~\cite{xie2024finben} & - & 0.17  & 0.00 & - & 0.41 & 0.00 & - & 0.17 & 0.00 & - & 0.17 & \textbf{0.00} & - & 0.17 & 0.00 & - & 0.17 & 0.00 & - & 0.17 & 0.00 & - & 0.17 & 0.00 & - & 0.17 & 0.00\\
LLaMA3-8B~\cite{xie2024finben} & - & 0.56 & \textbf{0.05} & - & 0.26 & 0.00 & - & 0.10 & -0.15 & - & 0.01 & \textbf{0.00} & - & 0.36 & - 0.03 & - & 0.83 & -0.06 & - & 0.26 & -0.07 & - & 0.94 & -0.01 & - & 0.00 & 0.00\\
\hline
\multicolumn{28}{c}{\textit{Few-shot LLM}} \\
GPT-4o~\cite{achiam2023gpt} & 0.61 & 0.73 & 0.53 & 0.86 & 0.84 & 0.51 & 0.62 & 0.51 & 0.65 & 0.59 & 0.65 & 0.73 & 0.51 & 0.58 & 0.61 & 0.58 & 0.51 & 0.67 & 0.73 & 0.66 & 0.62 & 0.45 & 0.57 & 0.52 & 0.60 & 0.58 & 0.72 \\

Deepseek V3~\cite{liu2024deepseek} & 0.68 & 0.57 & 0.58 & 0.59 & 0.59 & 0.53 & 0.55 & 0.65 & 0.53 & 0.50 & 0.65 & 0.56 & 0.68 & 0.62 & 0.68 & 0.47 & 0.51 & 0.71 & 0.50 & 0.48 & 0.71 & 0.54 & 0.64 & 0.65 & 0.51 & 0.72 & 0.69 \\

Qwen2.5 72B~\cite{hui2024qwen2} & 0.60 & 0.65 & 0.66 & 0.59 & 0.57 & 0.57 & 0.65 & 0.58 & 0.45 & 0.45 & 0.50 & 0.51 & 0.62 & 0.59 & 0.66 & 0.46 & 0.48 & 0.65 & 0.54 & 0.71 & 0.51 & 0.66 & 0.63 & 0.47 & 0.65 & 0.65 & 0.51 \\
\hline
\multicolumn{28}{c}{\textit{Financial LLM}} \\
FinMA-7B~\cite{xie2024finben} & - & 0.17 & 0.00 & - & - & 0.00 & - & 0.61 & 0.00 & - & 0.00 & \textbf{0.00} & - & 0.01 & -0.06 & - & 0.92 & -0.01 & - & 0.95 & 0.00 & - & 0.04 & 0.01 & - & 0.00 & 0.00\\
FinGPT-7B~\cite{xie2024finben} & - & 0.52 & 0.00 & - & 0.38 & 0.11 & - & 0.00 & 0.00 & - & \textbf{1.00} & \textbf{0.00} & - & 0.00  & 0.00 & - & 0.30 & 0.00 & - & 0.60 & -0.02 & - & \textbf{0.96} & 0.00 & - & \textbf{0.98} & 0.00\\
FinMA-ES-B~\cite{zhang2024dolares} & 0.60 & 0.60  & - & 0.72 & 0.71  & - & - & - & - & - & - & - & - & - & - & - & - & - & - & - & - & - & - & - & - & - & -\\ 
FinMA-ES-S~\cite{zhang2024dolares} & 0.66 & 0.52  & - & 0.56 & 0.51  & - & - & - & - & - & - & - & - & - & - & - & - & - & - & - & - & - & - & - & - & - & -\\ 
\hline
\multicolumn{28}{c}{\textit{PEFT Baselines}} \\
$\mathcal{M}_1$ - DoRA~\cite{liu2024dora} & 0.68 & 0.56 & 0.00 & 0.81 & 0.82 & 0.68 & 0.93 & 0.93 & 0.87 & 0.95 & 0.93 & 0.00 & 0.89 & 0.88 & 0.00 & 0.90 & 0.87 & 0.00 & 0.91 & 0.89 & 0.00 & 0.94 & 0.90 & 0.01 & 0.93 & 0.94 & 0.00 \\

$\mathcal{M}_2$ - DoRA~\cite{liu2024dora} & 0.70 & 0.55 & 0.00 & 0.82 & 0.82 & 0.69 & 0.91 & 0.91 & 0.86 & 0.96 & 0.95 & 0.00 & 0.90 & 0.90 & 0.00 & 0.91 & 0.88 & \textbf{\textcolor{blue}{0.02}} & 0.93 & 0.91 & \textbf{\textcolor{blue}{0.02}} & 0.92 & 0.89 & 0.00 & 0.92 & 0.92 & 0.00 \\

$\mathcal{M}_3$ - DoRA~\cite{liu2024dora} & 0.68 & 0.55 & \textbf{\textcolor{blue}{0.02}} & 0.80 & 0.80 & 0.68 & 0.92 & 0.92 & 0.87 & 0.94 & 0.94 & 0.00 & 0.88 & 0.91 & 0.00 & 0.90 & 0.88 & 0.00 & 0.93 & 0.93 & 0.00 & 0.93 & 0.90 & 0.00 & 0.95 & 0.92 & 0.00 \\

$\mathcal{M}_1$ - AdaLoRA~\cite{zhang2023adalora} & 0.70 & 0.56 & 0.00 & 0.83 & 0.79 & 0.68 & 0.94 & 0.92 & 0.87 & 0.94 & 0.94 & \textbf{\textcolor{blue}{0.02}} & 0.92 & 0.90 & 0.00 & 0.92 & 0.86 & 0.00 & 0.93 & 0.91 & 0.00 & 0.93 & 0.91 & 0.00 & 0.93 & 0.95 & \textbf{\textcolor{blue}{0.02}} \\

$\mathcal{M}_2$ - AdaLoRA~\cite{zhang2023adalora} & 0.71 & 0.54 & 0.00 & 0.80 & 0.82 & 0.66 & 0.91 & 0.94 & 0.86 & 0.95 & 0.93 & 0.00 & 0.89 & 0.92 & 0.00 & 0.89 & 0.89 & 0.01 & 0.92 & 0.92 & 0.01 & 0.92 & 0.90 & 0.00 & 0.93 & 0.91 & 0.00 \\

$\mathcal{M}_3$ - AdaLoRA~\cite{zhang2023adalora} & 0.68 & 0.55 & \textbf{\textcolor{red}{0.01}} & 0.82 & 0.80 & 0.68 & 0.95 & 0.94 & 0.89 & 0.93 & 0.95 & 0.00 & 0.90 & 0.89 & 0.00 & 0.89 & 0.86 & 0.01 & 0.90 & 0.91 & 0.00 & 0.94 & 0.90 & \textbf{\textcolor{blue}{0.01}} & 0.96 & 0.92 & 0.00 \\
\hline
\multicolumn{28}{c}{\textit{Our Approach (LAET)}} \\
$\mathcal{M}_1$ - LAET(Ours) & \textbf{\textcolor{red}{0.74}} & 0.61 & 0.00 & \textbf{\textcolor{red}{0.88}} & \textbf{\textcolor{red}{0.88}} & \textbf{\textcolor{red}{0.77}} & \textbf{\textcolor{red}{0.98}} & \textbf{\textcolor{red}{0.98}} & \textbf{\textcolor{blue}{0.93}} & \textbf{\textcolor{red}{1.00}} & \textbf{\textcolor{red}{1.00}} & 0.00 & \textbf{\textcolor{blue}{0.96}} & \textbf{\textcolor{red}{0.95}} & \textbf{\textcolor{red}{0.50}} & \textbf{\textcolor{red}{0.97}} & \textbf{\textcolor{red}{0.95}} & 0.00 & \textbf{\textcolor{red}{0.97}} & \textbf{\textcolor{red}{0.96}} & 0.00 & \textbf{\textcolor{red}{0.97}} & \textbf{\textcolor{blue}{0.95}} & 0.00 & \textbf{\textcolor{red}{0.99}} & \textbf{\textcolor{red}{0.98}} & 0.00 \\

$\mathcal{M}_2$ - LAET (Ours) & \textbf{\textcolor{red}{0.74}} & \textbf{\textcolor{blue}{0.62}} & 0.00 & 0.87 & 0.86 & 0.73 & \textbf{\textcolor{red}{0.98}} & \textbf{\textcolor{red}{0.98}} & \textbf{\textcolor{blue}{0.93}} & \textbf{\textcolor{red}{1.00}} & \textbf{\textcolor{red}{1.00}} & 0.00 & 0.95 & \textbf{\textcolor{red}{0.95}} & 0.02 & \textbf{\textcolor{red}{0.97}} & \textbf{\textcolor{blue}{0.94}} & 0.00 & \textbf{\textcolor{red}{0.97}} & \textbf{\textcolor{red}{0.96}} & 0.00 & \textbf{\textcolor{red}{0.97}} & \textbf{\textcolor{blue}{0.95}} & 0.00 & \textbf{\textcolor{red}{0.99}} & \textbf{\textcolor{red}{0.98}} & 0.00 \\

$\mathcal{M}_3$ - LAET (Ours) & \textbf{\textcolor{red}{0.74}} & \textbf{\textcolor{red}{0.63}} & 0.00 & \textbf{\textcolor{blue}{0.87}} & \textbf{\textcolor{blue}{0.87}} & \textbf{\textcolor{blue}{0.74}} & \textbf{\textcolor{red}{0.98}} & \textbf{\textcolor{red}{0.98}} & \textbf{\textcolor{red}{0.94}} & \textbf{\textcolor{red}{1.00}} & \textbf{\textcolor{red}{1.00}} & 0.00 & \textbf{\textcolor{red}{0.96}} & \textbf{\textcolor{red}{0.95}} & \textbf{\textcolor{blue}{0.31}} & \textbf{\textcolor{blue}{0.95}} & 0.92 & 0.00 & \textbf{\textcolor{red}{0.97}} & \textbf{\textcolor{red}{0.96}} & 0.00 & \textbf{\textcolor{red}{0.97}} & \textbf{\textcolor{red}{0.96}} & 0.00 & \textbf{\textcolor{red}{0.99}} & \textbf{\textcolor{red}{0.98}} & 0.00 \\
\bottomrule

\end{tabular}
\end{adjustbox}
\end{table*}

\begin{table}[ht]
\centering
\small
\caption{Performance Comparison of LAET models on FO benchmarks. Highest performance is marked by \textbf{\textcolor{red}{Bold red}} and the second highest given by \textbf{\textcolor{blue}{bold blue}}. }
\label{tab: table_4}
\setlength{\tabcolsep}{1.7pt}
\begin{adjustbox}{max width=0.47\textwidth}
\begin{tabular}{l c c c c c c c c c}
\toprule
\textbf{Models} & \multicolumn{3}{c}{\textbf{BigData22}} & \multicolumn{3}{c}{\textbf{ACL18}} & \multicolumn{3}{c}{\textbf{CIKM18}}\\ 
\cmidrule(lr){2-4} \cmidrule(lr){5-7} \cmidrule(lr){8-10}
& \textbf{Acc} & \textbf{F1} & \textbf{MCC} & \textbf{Acc} & \textbf{F1} & \textbf{MCC} & \textbf{Acc} & \textbf{F1} & \textbf{MCC}\\
 \midrule

\multicolumn{10}{c}{\textit{General Purpose LLM}} \\
ChatGPT~\cite{zhang2024dolares} & 0.53 & - & -0.025 & 0.50 & - & 0.005 & 0.55 & - & 0.005\\ 
GPT-4~\cite{zhang2024dolares} & 0.54 & - & 0.03 & 0.52 & - & 0.020 & 0.57 & - & 0.020\\ 
Gemini~\cite{xie2024finben} & 0.55 & - & 0.04 & 0.52 & - & 0.04 & 0.54 & - & 0.02\\
LLaMA2-7B~\cite{zhang2024dolares} & 0.51 & - & 0.030 & 0.51 & - & 0.010 & 0.47 & - & -0.070\\ 
LLaMA2-70B~\cite{xie2024finben} & 0.47 & - & 0.00 & 0.51 & - & 0.01 & 0.49 & - & -0.07\\
LLaMA3-8B~\cite{xie2024finben} & 0.55 & - & 0.02 & 0.52 & - & 0.02 & 0.57 & - & 0.03\\
\hline
\multicolumn{10}{c}{\textit{Few-shot LLM}} \\
GPT-4o~\cite{achiam2023gpt} & 0.53 & \textbf{\textcolor{blue}{0.54}} & 0.46 & 0.48 & 0.48 & 0.019 & 0.45 & 0.49 & 0.50 \\ 

Deepseek V3~\cite{liu2024deepseek} & 0.51 & 0.50 & 0.49 & 0.45 & 0.51 & 0.017 & 0.50 & 0.51 & 0.49 \\ 

Qwen2.5 72B~\cite{hui2024qwen2} & 0.53 & 0.53 & 0.47 & 0.50 & 0.45 & 0.025 & 0.51 & 0.52 & 0.48 \\
\hline
\multicolumn{10}{c}{\textit{Financial LLM}} \\
FinMA-7B~\cite{xie2024finben} & 0.51 & - & 0.02 & 0.51 & - & 0.03 & 0.50 & - & 0.08\\
FinGPT-7Bl~\cite{xie2024finben} & 0.45 & - & 0.00 & 0.49 & - & 0.00 & 0.42 & - & 0.00\\
FinMA-30B~\cite{xie2023pixiu} & 0.47 & - & 0.04 & 0.49 & - & 0.00 & 0.43 & - & -0.05\\
FinMA-ESB~\cite{zhang2024dolares} & 0.48 & - & 0.100 & 0.49 & - & -0.080 & 0.42 & - & -0.040\\ 
FinMA-ESS~\cite{zhang2024dolares} & 0.57 & - & 0.110 & 0.50 & - & -0.010 & 0.55 & - & -0.040\\ 
\hline
\multicolumn{10}{c}{\textit{PEFT Baselines}} \\
$\mathcal{M}_1$ - LoRA~\cite{hu2022lora} & 0.48 & 0.490 & 0.00 & 0.43 & 0.38 & 0.00 & 0.49 & 0.38 & 0.00 \\ 
$\mathcal{M}_2$ - LoRA~\cite{hu2022lora} & 0.51 & 0.53 & 0.01 & 0.46 & 0.35 & 0.02 & 0.496 & 0.372 & 0.000 \\  
$\mathcal{M}_3$ - LoRA~\cite{hu2022lora} & 0.480 & 0.460 & 0.007 & 0.477 & 0.388 & 0.003 & 0.501 & 0.362 & 0.019 \\ 
$\mathcal{M}_1$ - DoRA~\cite{liu2024dora} & 0.510 & 0.357 & 0.000 & 0.466 & 0.351 & 0.015 & 0.510 & 0.380 & 0.000 \\ 
$\mathcal{M}_2$ - DoRA~\cite{liu2024dora} & 0.482 & 0.355 & 0.000 & 0.458 & 0.380 & 0.000 & 0.488 & 0.390 & 0.000 \\ 
$\mathcal{M}_3$ - DoRA~\cite{liu2024dora} & 0.496 & 0.334 & 0.000 & 0.479 & 0.375 & 0.000 & 0.519 & 0.380 & 0.000 \\ 
$\mathcal{M}_1$ - AdaLoRA~\cite{zhang2023adalora} & 0.508 & 0.360 & 0.015 & 0.483 & 0.361 & 0.001 & 0.487 & 0.375 & 0.000 \\ 
$\mathcal{M}_2$ - AdaLoRA~\cite{zhang2023adalora} & 0.494 & 0.354 & 0.000 & 0.466 & 0.369 & 0.014 & 0.485 & 0.395 & 0.000 \\ 
$\mathcal{M}_3$ - AdaLoRA~\cite{zhang2023adalora} & 0.486 & 0.350 & 0.000 & 0.487 & 0.359 & 0.000 & 0.500 & 0.374 & 0.016 \\ 
\hline
\multicolumn{10}{c}{\textit{Our Approach (LAET)}} \\
$\mathcal{M}_1$ - LAET (Ours) & 0.550 & 0.400 & 0.000 & 0.520 & 0.420 & \textbf{\textcolor{blue}{0.017}} & \textbf{\textcolor{red}{0.580}} & 0.430 & 0.000 \\ 
$\mathcal{M}_2$ - LAET (Ours)& \textbf{\textcolor{red}{0.560}} & 0.460 & \textbf{\textcolor{red}{0.060}} & \textbf{\textcolor{red}{0.550}} & \textbf{\textcolor{blue}{0.520}} & \textbf{\textcolor{red}{0.097}} & 0.550 & \textbf{\textcolor{red}{0.550}} & \textbf{\textcolor{red}{0.105}} \\ 
$\mathcal{M}_3$ - LAET (Ours)& \textbf{\textcolor{blue}{0.570}} & \textbf{\textcolor{red}{0.550}} & \textbf{\textcolor{blue}{0.100}} & \textbf{\textcolor{blue}{0.530}} & \textbf{\textcolor{red}{0.510}} & 0.000 & \textbf{\textcolor{blue}{0.580}} & \textbf{\textcolor{blue}{0.560}} & \textbf{\textcolor{blue}{0.124}} \\ 
\bottomrule
\end{tabular}
\end{adjustbox}
\end{table}
\section{Experiment}
\subsection{Setup}
\noindent\textbf{Dataset:} We utilized 23 datasets covering 11 tasks, sourced from established benchmarks such as PIXIU~\cite{xie2023pixiu}, FinBen~\cite{xie2024finben}, and FLARE-ES~\cite{zhang2024dolares}. Each dataset $\mathcal{D} = \{(x_i, y_i)\}$ consists of query-answer pairs, where queries combine human-written instructions with task-specific inputs. 
\begin{tcolorbox}[colframe=gray!50, colback=gray!10, boxrule=0.2mm]
\scriptsize
\textbf{Input Text:} [instruction][text] "Answer:" \quad
\textbf{Response:} [answer]
\end{tcolorbox}
\noindent This unified format enables consistent instruction tuning across domains. While most labels are categorical (e.g., sentiment or classification tags), some datasets (e.g., TSA~\cite{pan2023evaluation}) use scalar outputs. This structure standardizes evaluation for a wide range of tasks beyond any single domain.

\noindent\textbf{Model Selection: }For our experiment, we focused on utilizing smaller LLMs and a neural network. Our goal was to ensure efficient performance in financial tasks with compact models, avoiding the need for larger models. Therefore, we selected LLMs with approximately 3 billion parameters and designed a lightweight neural network to facilitate layer-wise adaptation, supporting the overall pipeline for efficient task-specific tuning. \textbf{\textit{(a)} LLMs: }For our experiments, we considered three instruct tuned smaller LLMs: $\mathcal{M}_1$: \texttt{gemma-2-2b-it\footnote{https://huggingface.co/google/gemma-2-2b-it}} (2B parameters), $\mathcal{M}_2$: \texttt{Llama-3.2-3B-Instruct\footnote{https://huggingface.co/meta-llama/Llama-3.2-3B-Instruct}} (3.2B parameters). $\mathcal{M}_3$: \texttt{Phi-3.5-mini-instruct\footnote{https://huggingface.co/microsoft/Phi-3.5-mini-instruct}} (3.8B parameters). Although $\mathcal{M}_2$ and $\mathcal{M}_3$ support up to 128K token context windows and $\mathcal{M}_1$ supports 8K, we limit context length in practice to $\mathcal{C}_l \in [400, 4000]$ based on dataset size and computational efficiency. Complementing these models, \textbf{\textit{(b)} the shared classifier}  $\mathcal{F}$ consists of three fully connected layers: the input is first transformed by $z_1 = W_1 x + b_1$ into a 128-dimensional space, followed by a ReLU activation. This is passed through a second layer $z_2 = W_2 a_1 + b_2$ yielding a 64-dimensional representation, again followed by ReLU. Finally, the output is generated via $\hat{y} = W_3 a_2 + b_3$, where the output dimension corresponds to the number of task classes. For the regression-based TSA~\cite{pan2023evaluation} task, the output layer is modified to produce a scalar value with no activation, enabling direct prediction for MSE loss. This consistent and modular setup supports efficient layer-wise adaptation across tasks.
\noindent\textbf{Layer Probing Metrics: } During the process of selecting the best-performing layers, the function $\mathcal{F}$ is trained on the dataset $\mathcal{R}_l$, while the model $\mathcal{M}$ is kept in inference mode to prevent updates to the model's parameters. The performance of the layers is evaluated using two key metrics: accuracy ($m_1$) and F1-score ($m_2$). The hyperparameters $\alpha$ and $\beta$ are set to 0.5, ensuring equal weighting between the two performance measures. Additionally, the margin for $m_1$ and $m_2$ is defined as half of their respective standard deviations across the layers.

\noindent\textbf{Implementation Details\footnote{Codes and Data: TBA}. }The datasets, $\mathcal{D}$ used in this study are sourced from \texttt{The FinAI\footnote{https://huggingface.co/TheFinAI}} benchmark, as referenced in~\cite{xie2023pixiu, xie2024finben, zhang2024dolares}. We utilized \texttt{AutoModelForCausalLM} and \texttt{AutoTokenizer} from the \textit{Hugging Face Transformers} library to load all models, $\mathcal{M}$ and their respective tokenizers, $\mathcal{T}$. The small shared neural network, $\mathcal{F}$ was implemented using the \texttt{torch.nn} module. $\mathcal{F}$ was trained for $200$ epochs with a learning rate of $2e-4$ to determine the optimal layers. In the final training phase, the $\mathcal{M}$ and $\mathcal{F}$ were trained for 50 epochs. Backpropagation was applied to the unfrozen layers of the $\mathcal{M}$ with a learning rate of $2e-5$ and weight decay of $1e-4$, while the classifier was trained with a learning rate of $2e-4$. The pipeline for training is given in Algorithm~\ref{alg:alg_1}. The entire experiment was conducted using the \textit{PyTorch} framework on an \textit{RTX 6000 Ada} GPU.

\noindent\textbf{Evaluation Metrics: }We used Accuracy, F1 Score, and Matthews Correlation Coefficient (MCC) for evaluation. Table~\ref{tab:table_1} provides the specific evaluation metrics used for each dataset. Accuracy measures the overall proportion of correct predictions. The F1 Score, including micro and average F1, balances precision and recall to handle class imbalance, with micro F1 considering all instances equally and macro F1 averaging scores across classes. MCC provides a balanced evaluation by considering true and false positives and negatives, particularly effective in imbalanced datasets.

\subsection{Results and Analysis} 
\noindent We evaluate LAET's task-specific performance across Textual Analysis (TA), Forecasting (FO), and Risk Management (RM), benchmarking against state-of-the-art models. Additionally, we analyze model representations to assess their effectiveness for financial NLP tasks. \textbf{\textit{(1)} Textual Analysis.} As shown in Table~\ref{tab: table_2}, LAET outperforms existing benchmarks across 11 datasets spanning six TA tasks. Despite using smaller models, LAET variants ($\mathcal{M}_1$, $\mathcal{M}_2$, $\mathcal{M}_3$) consistently deliver top performance. Notably, $\mathcal{M}_3$ achieves 0.89 accuracy/F1 on FPB, 0.90 on FiQA, and a lowest RMSE of 0.18 on TSA. LAET also surpasses large models like GPT-4 and FinMA-7B on datasets such as FinanceES, Headlines, and FOMC, highlighting its robustness and efficiency. \textbf{\textit{(2)} Forecasting. }Table~\ref{tab: table_4} reports results on BigData22, ACL18, and CIKM18. LAET models offer competitive performance despite their smaller size. For example, $\mathcal{M}_3$ achieves 0.59 accuracy and 0.45 F1 on CIKM18, with the highest MCC. $\mathcal{M}_2$ scores 0.53 accuracy and 0.44 F1 on ACL18. These results confirm that compact, well-adapted models can rival larger LLMs in FO tasks. \textbf{Risk Management.} Despite class imbalance in RM datasets, LAET demonstrates strong adaptability (Table~\ref{tab: table_3}). $\mathcal{M}_1$ achieves 0.74 accuracy on German, $\mathcal{M}_2$ reaches 0.99 accuracy and 0.97 F1 on Polish, while $\mathcal{M}_3$ excels on LendingClub and Taiwan. LAET balances efficiency and performance, outperforming larger models without high compute costs.

\noindent \textbf{Observations: }Our experiments provided the following key insights: 
\textit{\textbf{(a)}} In textual analysis tasks, the number of effective layers increases with the complexity of classification. For binary classification (e.g., "yes" or "no"), earlier layers capture the necessary information, whereas multiclass problems benefit from deeper layers. 
\textit{\textbf{(b)}} In forecasting tasks, the nature of the data (e.g., time series) impacts LLM representations. Probing accuracy across all layers often remains around 50\%, indicating difficulty in capturing meaningful context from tokenized sequences. 
\textit{\textbf{(c)}} In highly imbalanced RM datasets (e.g., 99\%-1\% class splits), layers tend to become biased, with uniform probing accuracy across depths, making class differentiation challenging. 
\textit{\textbf{(d)}} LAET improves efficiency by freezing non-informative layers, enabling up to 60\% layer reduction in TA tasks without performance loss. 
\textit{\textbf{(e)}} LAET consistently outperforms existing PEFT methods such as LoRA, DoRA, and AdaLoRA, and surpasses large-scale models like GPT-4o, Deepseek V3, and Qwen2.5 72B—despite having significantly fewer parameters.

\begin{figure*}[!t]
    \centering    \includegraphics[width=\textwidth]{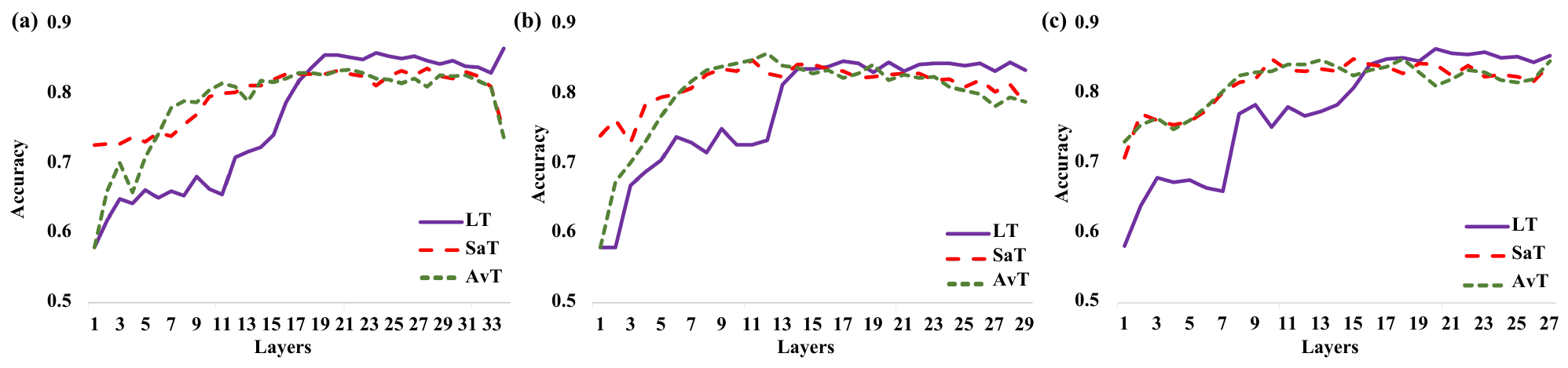}
    \caption{Layer-wise probing: We evaluated three probing strategies: Last Token (LT), Sum of All Tokens (SaT), and Average of All Tokens (AvT). The results indicate that LT consistently outperforms both SaT and AvT in accuracy, making it the most reliable indicator of layer effectiveness. This observation motivates our choice of the LT strategy in the LAET implementation.}

    \label{fig_3}
\end{figure*}

\begin{table}[!t]
\scriptsize
\centering
\caption{Performance comparison of different model configurations with varying $\alpha$ and $\beta$ values, number of layers, and parameters. Accuracy (Acc) and F1-score (F1) highlight the best-performing configuration ($\alpha = 0.5$, $\beta = 0.5$) in bold.}
\label{tab: table_5}

\scalebox{1.2}{%
\renewcommand{\arraystretch}{1.0}
\begin{tabular}{llllll}
\toprule
\textbf{$\alpha$} & \textbf{$\beta$} & \textbf{Layers} & \textbf{\# of Parameters} & \textbf{Acc} & \textbf{F1} \\ 
\midrule
0.3 & 0.3 & 15 & 1782939011 & 0.85 & 0.84\\ 

0.3 & 0.7 & 18 & 2122696067 & 0.88 & 0.87\\ 

0.7 & 0.7 & 18 & 2122696067 & 0.88 & 0.87\\ 

0.7 & 0.3 & 18 & 2122696067 & 0.87 & 0.86\\ 

0.5 & 0.5 & 17 & 2009443715 & \textbf{0.89 }& \textbf{0.88}\\ 
\bottomrule
\end{tabular}
}
\end{table}
\noindent\textbf{Hidden State Representation Analysis: }
Unlike BERT-based models, which have a dedicated classifier token (like `[CLS]`), LLMs do not have a specific token for classification. To address this, we analyze hidden representations using three strategies: summing all tokens' representations, averaging them, and using the last token's hidden state. Our results show that the last token provides the best probing accuracy, as it captures the most relevant information due to the model's sequential attention mechanism. This performance is visualized in Fig.~\ref{fig_3}.

\subsection{Ablation Study}
\begin{figure}[!t]
    \centering
    \includegraphics[width=0.4\textwidth]{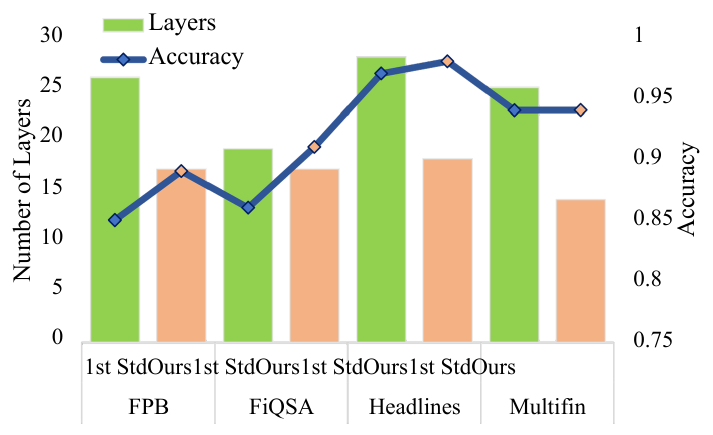}
    \caption{The 1st Std method selects 25–30 layers, while the proposed method selects 15–22, reducing layers by 20–40\% while maintaining accuracy above 0.9, proving its efficiency.}
    \label{fig_4}
\end{figure}

\noindent \textbf{Layer Selection formulation: } One of the most common layer selection criteria is based on standard deviation. Fig.~\ref{fig_4} shows the comparison of layer selection based on the first standard deviation and our proposed method  across four datasets. The standard deviation-based approach selects more layers, while the proposed method selects fewer. Despite using fewer layers, the proposed method achieves similar or better accuracy, demonstrating its efficiency. This suggests that selecting layers based on the proposed method leads to a more compact model without compromising performance.

\noindent \textbf{Domain-Agnostic LAET:} Table~\ref{tab: table_6} presents results on two benchmarks: the ETHICS dataset for moral reasoning and the Medical Tweets dataset for domain-specific classification. While GPT-4o provides a strong baseline, LAET consistently surpasses GPT-4o, base models, and PEFT variants, demonstrating its robustness and applicability beyond finance.

\noindent \textbf{Impact of Margin Parameter ($\alpha$, $\beta$): } We have examined the effect of the parameters $\alpha$ and $\beta$ on layer selection and optimization. Our findings (showed in table~\ref{tab: table_5}) indicate that the chosen values for $\alpha$ and $\beta$ yielded the most optimal trade-off between parameter efficiency and accuracy.

\begin{table}[!t]
\small
\centering
\caption{Performance comparison across Ethics and Medical data. 
Best values in \textbf{\textcolor{red}{red bold}}, second-best in \textbf{\textcolor{blue}{blue bold}}.}
\label{tab: table_6}
\renewcommand{\arraystretch}{1.}
\setlength{\tabcolsep}{8.5pt}
\begin{tabular}{l cc cc}
\hline
\multirow{2}{*}{\textbf{Model}} & 
\multicolumn{2}{c}{\textbf{Ethics~\cite{hendrycks2020aligning}}} & 
\multicolumn{2}{c}{\textbf{Medical~\cite{jayasurya2021analysis}}} \\
\cline{2-5}
 & \textbf{Acc} & \textbf{F1} & \textbf{Acc} & \textbf{F1} \\
\hline
GPT-4o~\cite{achiam2023gpt}    & 0.67 & 0.66 & 0.61 & 0.62 \\
\hline
\multicolumn{5}{c}{\textit{Base Models}} \\
$\mathcal{M}_1$     & 0.47 & 0.45 & 0.53 & 0.55 \\
$\mathcal{M}_2$     & 0.53 & 0.54 & 0.48 & 0.50 \\
$\mathcal{M}_3$     & 0.56 & 0.56 & 0.52 & 0.53 \\
\hline
\multicolumn{5}{c}{\textit{PEFT Baselines}} \\
$\mathcal{M}_1$-LoRA~\cite{hu2022lora}       & 0.61 & 0.61 & 0.58 & 0.60 \\
$\mathcal{M}_2$-LoRA~\cite{hu2022lora}       & 0.64 & 0.66 & 0.69 & 0.67 \\
$\mathcal{M}_3$-LoRA~\cite{hu2022lora}       & 0.64 & 0.65 & 0.67 & 0.69 \\
$\mathcal{M}_1$-DoRA~\cite{liu2024dora}      & 0.62 & 0.62 & 0.60 & 0.59 \\
$\mathcal{M}_2$-DoRA~\cite{liu2024dora}      & 0.65 & 0.63 & 0.70 & 0.69 \\
$\mathcal{M}_3$-DoRA~\cite{liu2024dora}      & 0.66 & 0.66 & 0.72 & 0.71 \\
$\mathcal{M}_1$-AdaLoRA~\cite{zhang2023adalora}  & 0.61 & 0.61 & 0.58 & 0.60 \\
$\mathcal{M}_2$-AdaLoRA~\cite{zhang2023adalora}  & 0.64 & 0.66 & 0.69 & 0.67 \\
$\mathcal{M}_3$-AdaLoRA~\cite{zhang2023adalora}  & 0.64 & 0.65 & 0.67 & 0.69 \\
\hline
\multicolumn{5}{c}{\textit{Our Approach (LAET)}} \\
$\mathcal{M}_1$-LAET (Ours)    & 0.68 & 0.67 & 0.74 & 0.76 \\
$\mathcal{M}_2$-LAET (Ours)    & \textbf{\textcolor{blue}{0.70}} & \textbf{\textcolor{blue}{0.71}} & \textbf{\textcolor{blue}{0.78}} & \textbf{\textcolor{blue}{0.78}} \\
$\mathcal{M}_3$-LAET (Ours)    & \textbf{\textcolor{red}{0.76}} & \textbf{\textcolor{red}{0.74}} & \textbf{\textcolor{red}{0.84}} & \textbf{\textcolor{red}{0.86}} \\
\hline
\end{tabular}
\end{table}

\subsection{Discussion}

\noindent Our experiments demonstrate that lightweight LLMs, such as $\mathcal{M}_1$, $\mathcal{M}_2$, and $\mathcal{M}_3$, achieve competitive performance in financial tasks like textual analysis, forecasting, and risk management, often outperforming larger models like GPT-4. Layer-wise adaptation and fine-tuning proved effective, allowing up to 60\% of layers to be frozen in textual tasks without performance loss, significantly improving efficiency. However, forecasting tasks revealed limitations, with LLMs struggling to extract meaningful patterns from time-series data, while risk management tasks highlighted challenges with class imbalance. The last token's hidden state emerged as the most effective representation for classification, leveraging the sequential attention mechanism of LLMs. These findings underscore the potential of lightweight models for financial NLP tasks while identifying areas for further improvement, such as hybrid approaches for forecasting and techniques to address class imbalance.

\noindent \textbf{Limitations:}
Our study primarily focused on single-word classification using lightweight LLMs, demonstrating their effectiveness in financial tasks like textual analysis and classification. However, the approach could be extended to a wider range of tasks, such as Named Entity Recognition (NER), Question Answering (QA), and Summarization, through more comprehensive pipelines like LAET. The lack of large annotated datasets limited our ability to conduct experiments on a larger scale and explore these tasks in-depth. Moreover, while lightweight models performed well, larger models with more parameters may offer improved task-specific performance, especially with better data and task adaptation strategies.

\section{Conclusion}
\noindent In this study, we have proposed a novel strategy LAET, that optimize models by layer wise probing. We implemented the pipeline for specific financial NLP tasks, demonstrating that smaller models like can achieve competitive performance across textual analysis, forecasting, and risk management tasks. Our layer-wise adaptation and fine-tuning strategy proved highly efficient, enabling significant computational savings by freezing ineffective layers without sacrificing accuracy. While these models excelled in textual analysis and risk management, forecasting tasks revealed limitations, suggesting the need for hybrid approaches to better handle time-series data. Additionally, the last token's hidden state emerged as the most effective representation for classification tasks, leveraging the sequential nature of LLMs. These findings highlight the potential of lightweight models for financial applications, offering a balance between performance and efficiency. Future work could focus on addressing challenges in forecasting and class imbalance, as well as exploring hybrid architectures to further enhance model capabilities in financial NLP tasks.

\noindent \textbf{Future Works:}
Future research will focus on extending the application of lightweight LLMs to a broader range of financial tasks, including NER, QA, and Summarization, leveraging frameworks like the LAET pipeline. Additionally, addressing the limitations observed in time-series forecasting tasks, such as improving the model's ability to extract meaningful patterns from temporal data, will be a priority. We also aim to explore hybrid model approaches to enhance forecasting accuracy and develop techniques to tackle class imbalance in risk management tasks. Furthermore, expanding experiments on larger-scale annotated datasets will help assess the scalability and generalizability of our models across various domains.

\section*{Ethics Statement}
\noindent Our research on financial sentiment analysis using lightweight LLMs adheres to ethical standards, ensuring fairness, transparency, and integrity. We used publicly available, anonymized datasets to protect privacy and avoided the use of proprietary or sensitive financial data without authorization. While advancing AI technology for financial decision-making, we acknowledge the potential risks and are committed to addressing any biases or limitations in our models, promoting responsible and fair deployment in real-world applications.


\bibliographystyle{IEEEtran}
\bibliography{reference}

\end{document}